# On Inconsistency Indices and Inconsistency Axioms in Pairwise Comparisons

Jiří Mazurek
*Silesian University in Opava, Czech Republic*
mazurek@opf.slu.cz

**Abstract**: *Pairwise comparisons are an important tool of modern (multiple criteria) decision making. Since human judgments are often inconsistent, many studies focused on the ways how to express and measure this inconsistency, and several inconsistency indices were proposed as an alternative to Saaty's inconsistency index and inconsistency ratio for reciprocal pairwise comparisons matrices. This paper aims to: firstly, introduce a new measure of inconsistency of pairwise comparisons and to prove its basic properties; secondly, to postulate an additional axiom, an upper boundary axiom, to an existing set of axioms; and the last, but not least, the paper provides proofs of satisfaction of this additional axiom by selected inconsistency indices as well as it provides their numerical comparison.*

## Introduction

Pairwise comparisons are a well-known decision making method with history dating back to early works of Lull, Condorcet and Thurstone's *A Law of Comparative Judgments*, see Thurstone (1927). They enable to compare two objects, usually alternatives, criteria, qualities or features, at the same time. Pairwise comparisons are especially useful when the number of object to be compared is large, as they reduce complexity of a problem and aid to avoid a cognitive overload, as according to Miller (1956), humans are only able to compare 7 objects at one time. Also, from pairwise comparisons a priority vector (a vector of weights of compared objects) can be derived via some well-known method such as the eigenvalue method or the geometric mean method (the least logarithmic squares method), for the former see e. g. see Saaty (1980, 2004 and 2008), for the latter see e. g. Chandran et al. (2005). In particular, the analytic hierarchy process (AHP) is perhaps the best known application of pairwise comparisons, successful applications of the AHP can be found e. g. in Vaidya and Kumar (2006).

Recently, research on pairwise comparisons is focused on the problem of inconsistency of pairwise comparisons and their comparison and axiomatization. When comparing for example three alternatives, A, B and C, a decision maker has to perform three pairwise comparisons: A with B, A with C and B with C. If A is set to be 2 times better than B and B is set to be 3 times better than C, A should be 6 times better than C. Although the task seems simple, decision makers are seldom consistent, and their judgments are "erroneous" to a certain degree. To measure the inconsistence of their judgments, several inconsistency indices (see section 2) were proposed.

Nevertheless, these indices can be considered ad-hoc in a sort of way; they were introduced independently and spontaneously. That is why the problem of axiomatization is studied extensively in the last years, see Koczkodaj and Szwarc (2014), Ramík (2015), Brunelli and Fedrizzi (2015) or Csató (2016). Nevertheless, the problem of axiomatization is not over yet, as new indices emerge, especially in extensions of pairwise comparisons to fuzzy sets and Abelian linearly ordered groups, see for instance Cavallo and D'Apuzzo (2009) or Ramík and Korvíny (2010).

This paper aims to contribute to the ongoing research. The main aim of this paper is to introduce a new inconsistency index, which is based on a dot (scalar) product of row vectors of a pairwise comparisons matrix, and to prove its properties. Also, properties of Koczkodaj's inconsistency index are proved as well. Last, but not least, another sixth axiom, the upper boundary for index's value, is added to the five axioms of Brunelli and Fedrizzi (2015), and it is examined whether selected inconsistency indices satisfy it or not.

The paper is organized as follows: in section 1 preliminaries are provided, section 2 describes inconsistency indices and section 3 inconsistency axioms, in section 4 a new inconsistency index is introduced and its properties are proved, in section 5 selected indices are examined with respect to an additional axiom, and section 6 includes a numerical example. Conclusions close the article.

## 1. Preliminaries

For simplicity, but without loss of generality, let us consider pairwise comparisons of alternatives.

Let $X$ be a given set of $n$ alternatives to be compared. Let $a_{ij}$ denote the preference of $i$-th alternative over $j$-th alternative. Also, we set $a_{ij} > 0; \forall i, j \in \{1, 2, ..., n\}$.

Pairwise comparisons are called reciprocal, if the following property is satisfied:

$$a_{ij} = \frac{1}{a_{ji}}, \forall i, j \in \{1, 2, ..., n\}. \tag{1}$$

The property (1) is usually strictly required for pairwise comparisons. All pairwise comparisons can be arranged into a square $n \times n$ matrix $A(a_{ij})$ called a pairwise comparison matrix (PCM):

$$A_{n \times n}(a_{ij}) = \begin{pmatrix} 1 & a_{12} & ... & a_{1n} \\ a_{21} & 1 & ... & a_{2n} \\ ... & ... & 1 & ... \\ a_{n1} & a_{n2} & ... & 1 \end{pmatrix}$$

Further, pairwise comparisons are called consistent, if the following property is satisfied:

$$a_{ij} \cdot a_{jk} = a_{ik}; \forall i, j, k. \tag{2}$$

From a pairwise comparisons matrix $A$ weights of alternatives (a priority vector) $w$ can be derived via Saaty's eigenvalue method (EM):

$$Aw = \lambda_{max} w, \tag{3}$$

where $\lambda_{max}$ is the largest (positive) eigenvalue of $A$. The existence of the largest (positive) eigenvalue $\lambda_{max}$ of the matrix $A$ is guaranteed by Perron-Frobenius theorem (Saaty 1980).

Also, the geometric mean method (the least squares method) can be used to obtain $w$:

$$w_i = \frac{\left(\prod_{j=1}^{n} a_{ij}\right)^{1/n}}{\sum_{i=1}^{n}\left(\prod_{j=1}^{n} a_{ij}\right)^{1/n}} \quad (4)$$

Both methods yield the same result when the matrix $A$ is consistent. Otherwise, the priority vectors differ slightly, see the comparative study of Ishizaka and Lusti (2006).

For further considerations it should be noted that if the matrix $A$ is reciprocal and consistent, then $\lambda_{max} = n$ and rank$(A) = 1$, which means all rows (columns) of $A$ differ only by a multiplicative constant.

***Definition 1.*** Let $M_R$ denote the set of all matrices $A_{n \times n}(a_{ij})$, $a_{ij} > 0; \forall i, j \in \{1, 2, ..., n\}$, satisfying (1), and let $M_C$ denote the set of all matrices $A_{n \times n}(a_{ij})$, $a_{ij} > 0; \forall i, j \in \{1, 2, ..., n\}$ satisfying (2).

By the definition above, $M_R$ is the set of all reciprocal pairwise comparison matrices with positive elements of the order $n$, and $M_C$ is the set of all consistent pairwise comparison matrices with positive elements of the order $n$.

***Remark 1.*** As the matrix $A$ with elements $a_{ij} = 1, \forall i, j$ belongs both to $M_R$ and $M_C$, both sets are non-empty, and clearly: $M_C \subset M_R$. Furthermore, consistency (2) implies reciprocity (1), but not vice versa.

***Definition 2.*** A corner pairwise comparison matrix $CPC$ of the order $n$ is defined as follows:

$$CPC_{n \times n} = \begin{pmatrix} 1 & 1 & ... & 1 & x \\ 1 & 1 & ... & 1 & 1 \\ ... & ... & ... & ... & ... \\ 1 & 1 & ... & 1 & 1 \\ 1/x & 1 & ... & 1 & 1 \end{pmatrix}, x > 0.$$

A corner pairwise comparison matrix is a special (simplified) case of a reciprocal matrix, which is consistent if and only if $x = 1$.

## 2. Inconsistency indices

In this section, several inconsistency indices for a pairwise comparison matrix are going to be introduced, starting with the oldest one, Saaty's consistency index *C.I.*

- *Consistency index C.I.* (Saaty, 1980):

$$C.I. = \frac{\lambda_{max} - n}{n - 1} \quad (5)$$

According to Saaty, a PCM with *C.I.* lower or equal to 0.10 is sufficiently consistent, and EM method (3) can be employed. However, *C.I.* grows with $n$, that is why Saaty introduced a more suitable measure of inconsistency, the consistency ratio *C.R.*

- *Consistency ratio C. R.* (Saaty, 2004, 2008):

$$C.R. = \frac{C.I.}{R.I.} \quad (6)$$

The *R.I.* in (6) is the random inconsistency, that is an average of *C.I.* of random matrices (generated by the Monte Carlo method) of the order *n*. For the values of *R.I.* see e.g. (Alonso and Lamata, 2006).

Again, a PCM with *C.R.* lower or equal to 0.10 is sufficiently consistent for the eigenvalue method. Nevertheless, the rule of thumb of 0.10 was criticized by some authors, see Dyer (1990) or Koczkodaj and Szybowski (2014). Notably, *R.I.* was found to converge to the value 1.58 with increasing *n*. This fact drew some criticism of *C.R.* (and *C.I.*), as random inconsistency should be increasing with the increasing *n* (the larger is a random matrix, the more "mess" it contains).

- *Index GWI*, Golden and Wang (1989): Let $A(a_{ij}) \in M_R$ and let $\overline{A}(\overline{a}_{ij})$ be the normalized matrix obtained from *A* by dividing each column by a sum of all elements in a corresponding column. Further, let $\overline{w}$ be the normalized priority vector (the vector of weights) obtained from *A* by the EM or the geometric mean method. Then the inconsistency index *GWI* is defined as:

$$GWI = \frac{1}{n} \sum_{i=1}^{n} \sum_{j=1}^{n} |\overline{a}_{ij} - \overline{w}_i| \quad (7)$$

- *Index PLI*, Pelaez-Lamata (2003): Let $A(a_{ij}) \in M_R$. Then the inconsistency index *PLI* is defined as follows:

$$PLI = \frac{6}{n(n-1)(n-2)} \sum_{i=1}^{n-2} \sum_{j=i+1}^{n-1} \sum_{k=j+1}^{n} \left( \frac{a_{ik}}{a_{ij}a_{jk}} + \frac{a_{ij}a_{jk}}{a_{ik}} - 2 \right) \quad (8)$$

- *Geometric consistency index* (*GCI*), (Aguaron and Moreno-Jimenez, 2003):

$$GCI(A) = \frac{2}{(n-1)(n-2)} \sum_{i=1}^{n-1} \sum_{j=i+1}^{n} \ln^2 \left( a_{ij} \frac{w_j}{w_i} \right) \quad (9)$$

- *Koczkodaj's index (KII)*, Koczkodaj (1993, 2014): Let T(*n*) be the set of all ordered triples ("triads") $\left( a_{ij}, a_{jk}, a_{ik} \right)$ satisfying relation (2) for $\forall i,j,k \in \{1,2,...,n\}$. Then:

$$KII = \max_{T(n)} \left( \min \left( \left| 1 - \frac{a_{ik}}{a_{ij}a_{jk}} \right|, \left| 1 - \frac{a_{ij}a_{jk}}{a_{ik}} \right| \right) \right) \quad (10)$$

For other inconsistency indices, see e.g. Barziali (1998), Stein and Mizzi (2007), Ramík and Korvíny (2010), Ramík (2015).

Majority of indices capture a central tendency (a mean) of inconsistent judgments. On the other hand, Koczkodaj's inconsistency index aims to express the largest inconsistency present in a given PCM. It should be noted that indices such as *PLI* or *GWI* can be easily modified to return the maximum inconsistency as well, by substituting the averaging operators for the maximum operator. Therefore, indices might be divided into two groups:

- Mean-based (*C.I.*, *C.R.*, *GWI*, *PLI*, *GCI*),
- Extreme-based (*KII*).

Advantage of the mean-based indices is that they take into account every pairwise comparison (and its changes), but they do not provide (explicitly) information about extremes. On the other hand, extreme-based indices express the inconsistency in terms of the most inconsistent judgment, which can be useful when the most "erroneous" comparison is to be found and revised. Disadvantage of this class of indices is that all changes besides the most inconsistent comparison (a triad) are neglected.

This may lead to the idea of a new (compromise) family of indices of the following form:

$$IC_{compr} = \lambda \cdot IC_{mean} + (1-\lambda) \cdot IC_{extreme}, \lambda \in [0,1],$$

where a decision maker decides how he/she wants to capture both features by selecting an appropriate value of $\lambda$, where $\lambda = 0$ yields an extreme index and $\lambda = 1$ a central based index.

## 3. Axioms for inconsistency indices

An inconsistency index should have some reasonable properties. Koczkodaj and Szwarc (2014) introduced three axioms an inconsistency index must satisfy, while Brunelli and Fedrizzi (2015) postulated five axioms. Two axioms are identical in both papers.

Below, the five axioms according to Brunelli and Fedrizzi (2015) are provided with slight modifications, along with a newly formulated Axiom 6.

***Definition 3***. An inconsistency index (*ICI*) is a real-valued function:

$$ICI: A \in M_R \to [0, \infty],$$
or alternatively, $ICI: A \in M_R \to [0,1].$

***Axiom 1***. $A \in M_C \Leftrightarrow ICI(A) = 0$.

Axiom 1 states that consistent pairwise comparison matrices are identified by a unique real value of inconsistency. Usually, this value is set to 0, as zero inconsistency means consistency.

***Axiom 2***. Let *P* denote a permutation matrix of the order *n* and let $A_{n \times n} \in M_R$. Then for all *P* and all *A´*, so that $A´ = P \cdot A \cdot P^T$, the following condition holds:

$$ICI(A´) = ICI(A), \forall P.$$

Axiom 2 expresses that the inconsistency index is invariant under permutation of alternatives. In other words, changing of an order of alternatives should not result in a change of an inconsistency index.

***Axiom 3***. Let *f* be a continuous transformation: $f: a_{ij} \to a_{ij}^b; \forall i,j; b \in R, b > 1$. Let the matrix with elements $a_{ij}^b$ be denoted as $A^b$. Then:

$$ICI(A^b) \geq ICI(A)$$

Axiom 3 deals with monotonicity of intensity of preference: if preferences are intensified, then an inconsistency index cannot return a lower value.

***Axiom 4***. Let $A \in M_C$ and let at least one $a_{ij} \neq 1$ for $i \neq j$. Let $A^\delta$ denote a matrix obtained from $A$ by substituting the element $a_{ij} \neq 1$ (and also $a_{ji} \neq 1$) by the element $a_{ij}^\delta$, where $\delta \in R, \delta > 0, \delta \neq 1$. Then $ICI(A^\delta)$ is a non-decreasing function for $\delta > 1$ and a non-increasing function for $\delta < 1$.

Axiom 4 requires monotonicity on a single comparison: the larger is the change in a given entry $a_{ij}$ (and $a_{ji}$ respectively) from a consistent matrix, the more inconsistent is the resulting matrix and the higher is the value of an inconsistency index.

***Axiom 5***. An inconsistency index is a continuous function of its entries.

Axiom 5 ensures there are not "jumps" or other discontinuities of *ICI* values.
Whether the presented set of five axioms is the best possible (optimal) choice is certainly open to a discussion.
One feature not captured by the five axioms is the problem of upper boundary of an inconsistency index. If an inconsistency index is not bounded from above, it might be problematic to interpret its values (what information a decision maker obtains when *ICI* is, for example, equal to 984,669?). Therefore, it seems more natural to introduce an upper bound on *ICI*.

***Axiom 6***. *ICI* is bounded from above if and only if:

$$\exists K \in R; ICI(A) \leq K, \forall A \in M_R.$$

Table 1 summarizes which axioms are (not)satisfied by indices (5), (7-11). It is based on Brunelli and Fedrizzi (2015), but it is extended to encompass indices *KII* and *RIC* (introduced later), and also it is extended to Axiom 6.

Table 1. A satisfaction of axioms by inconsistency indices.

| Index/Axiom | A1 | A2 | A3 | A4 | A5 | A6 |
|:---:|:---:|:---:|:---:|:---:|:---:|:---:|
| C.I. | Y | Y | Y | Y | Y | N |
| GWI | Y | Y | N | ? | Y | Y |
| GCI | Y | Y | Y | Y | Y | N |
| PLI | Y | Y | Y | Y | Y | N |
| RIC | Y | Y | N | Y | Y | Y |
| KII | Y | Y | Y | Y | Y | Y |

Note: "Y" means an index satisfies an axiom; "N" means an index does not satisfy an axiom and "?" means the result is open.

## 4. A new index of inconsistency of a pairwise comparisons matrix

The concept of this new measure of inconsistency of a pairwise comparisons matrix (PCM) comes from a geometrical point of view: the rows of a PCM can be considered vectors in the $n$-dimensional Euclidean space. If a PCM is fully consistent, then the PCM's rows are collinear (they differ only by a multiplicative constant). When an inconsistency appears, the rows will not be collinear any more, and their "deviation" from collinearity can be expressed by a cosine of an angle between each pair of row vectors.

Again, we set $a_{ij} \in (0, \infty)$.

Let $r_i = (a_{i1}, ..., a_{in})$ and $r_j = (a_{j1}, ..., a_{jn})$ be the row vectors of a pairwise comparison matrix $A \in M_R$ of the order $n$. Then the inconsistence index based on the rows of $A$ is given as:

- *Row inconsistency index (RIC)*: Let $A_{n \times n}(a_{ij}) \in M_C$. Then *RIC* is given as follows:

$$RIC = 1 - \frac{2\sum_{i=1}^{n-1}\sum_{j>i}^{n} \cos\varphi_{ij}}{n(n-1)}, \qquad (11)$$

where $\cos\varphi_{ij} = \dfrac{r_i \cdot r_j}{|r_i| \cdot |r_j|}$, and $r_i \cdot r_j$ denotes the dot product. $\qquad (12)$

Index *RIC* is the only "geometrically" based inconsistency index, which is equal to 1 minus the arithmetic mean of cosines of all pairs of row vectors of a given PCM. Also, the values of this index are conveniently bounded to the interval $[0,1]$ (which is going to be proved below), unlike the indices PLI (8) and GCI (9). It should be noted that although the definition of *RIC* is based on the row vectors, the use of column vectors of the matrix $A$ would be feasible as well.

Now, some properties satisfied by RIC index (11) will be discussed.

**Remark 2.** Because $a_{ij} > 0, \forall i, j$, the angle $\varphi_{ij}$ satisfies $0 \leq \varphi_{ij} < 90°$ for all $r_i$ and $r_j$, and $0 \leq \cos\varphi_{ij} < 1$.

**Proposition 1.** $A \in M_C \Leftrightarrow RIC = 0$.

*Proof*: When $A$ is consistent, its rows are collinear, thus $\varphi_{ij} = 0, \forall i, j$ and $\cos\varphi_{ij} = 1, \forall i, j$, which yields $ICI = 0$.

If $RIC = 0$, then $\dfrac{2\sum_{i=1}^{n-1}\sum_{j>i}^{n} \cos\varphi_{ij}}{n(n-1)} = 1$. Since the codomain of cosine is $[-1,1]$ and the number of cosine terms in the nominator is $\binom{n}{2}$, we get immediately $\cos\varphi_{ij} = 1, \forall i, j > i$; hence, the row vectors $r_i$ are collinear, rank $(A) = 1$ and $A \in M_C$.

**Proposition 2.** $RIC > 0 \Leftrightarrow A$ is not consistent.

*Proof:* Assume that $RIC > 0$, then at least one $\cos\varphi_{ij} < 1$, which means at least one $\varphi_{ij} > 0$, therefore there are at least two row vector that are non-collinear; hence the rank $(A) \geq 2$ and $A$ is not consistent.

Now assume that $A$ is not consistent. Then rank $(A) \geq 2$, so at least two row vectors are not collinear and at least one $\varphi_{ij} > 0$, therefore at least one element $\cos\varphi_{ij} < 1$, which yields $RIC > 0$.

**Proposition 3.** $0 \leq RIC \leq 1$.

*Proof:* Since all elements of a matrix $A$ are positive, all row vectors $r$ lie in the first orthant of the space $R^n$. Hence, from (12) we get $\cos\varphi_{ij} \geq 0, \forall i, j > i$, but also $\cos\varphi_{ij} \leq 1, \forall i, j > i$.

Therefore, $0 \leq \dfrac{2\sum_{i=1}^{n-1}\sum_{j>i}^{n}\cos\varphi_{ij}}{n(n-1)} \leq 1$, and finally $0 \leq RIC = 1 - \dfrac{2\sum_{i=1}^{n-1}\sum_{j>i}^{n}\cos\varphi_{ij}}{n(n-1)} \leq 1$.

**Proposition 4.** *RIC is an increasing function in each argument $\varphi_{ij}$.*

*Proof:* It suffices to show that the first derivative of (1) is positive:

$$(RIC)' = \dfrac{2\sum_i\sum_j \sin\phi_{ij}}{n(n-1)} \geq 0 \,\forall i,j, \text{ since } \sin\varphi_{ij} \geq 0 \text{ for all } \varphi_{ij} \in [0, 90].$$

**Proposition 5.** *RIC is a concave function in each argument $\varphi_{ij}$.*

*Proof:* Let fix all $\varphi_{ij}$ except for one. Then, a function of one (real) variable is concave if its second derivative is negative. Therefore, we get: $RIC = 1 - \dfrac{2\sum_i\sum_j \cos\phi_{ij}}{n(n-1)}$, so $(RIC)' = \dfrac{2\sum_i\sum_j \sin\phi_{ij}}{n(n-1)}$ and $(RIC)'' = -\dfrac{2\sum_i\sum_j \cos\phi_{ij}}{n(n-1)}$. Because $\cos\varphi_{ij} \geq 0, \forall i,j$ we obtain $(RIC)'' \leq 0$.

**Proposition 6.** *RIC satisfies Axiom 1.*
*Proof:* It was already proven in Proposition 1.

**Proposition 7.** *RIC satisfies Axiom 2.*
*Proof:* The statement is obvious, as the average value of $\cos\varphi_{ij}$ in (11) is computed from all pairs of rows regardless of their order (all rows are treated equally).

**Proposition 8.** *RIC does not satisfy Axiom 3.*
*Proof* by a counterexample: Let $A \in M_R$ be the pairwise comparisons matrix as follows:

$$A = \begin{pmatrix} 1 & 0.1 & 0.15 \\ 10 & 1 & 0.3 \\ 6.6666 & 3.3333 & 1 \end{pmatrix}, RIC(A) = 0.047.$$

Let $B$ be the pairwise comparisons matrix obtained from the matrix $A$ by raising each entry of the matrix $A$ to the power of two:

$$B = A^{(b=2)} = \begin{pmatrix} 1 & 0.01 & 0.0225 \\ 100 & 1 & 0.09 \\ 44.4444 & 11.1111 & 1 \end{pmatrix}, RIC(B)=0.018$$

Thus, although preferences expressed in the matrix $B$ are "stronger", the inconsistency index *RIC* returns a smaller value.

*Proposition 9*. *RIC* satisfies Axiom 4.
*Proof*: As *RIC* is increasing with the angle $\varphi$ between any two row vectors of $A$ (see Proposition 4), the more the matrix $A$ changes in one element (and its reciprocal) from a consistent form, the larger is the angle $\varphi$ between this row and other rows, hence the larger is *RIC* as well.

*Proposition 10*. *RIC* satisfies Axiom 5.
*Proof*: The proof is obvious as the cosine function in (11) is continuous.

*Proposition 11*. *RIC* satisfies Axiom 6.
*Proof*: *RIC* <1 was already proven in Proposition 3.

## 5. Koczkodaj's index and the satisfaction of the Axiom 6 by *C.I.*, *GWI*, *PLI*, and *GCI*

The *KII* index was omitted in Brunelli and Fedrizzi (2015) paper, but it is an important inconsistency index since it is the only extreme-based one. Therefore, now its properties are going to be proved.

*Proposition 12*. *KII* satisfies Axiom 1.
*Proof:* It is obvious from its definition.

*Proposition 13*. *KII* satisfies Axiom 2.
*Proof:* Since *KII* evaluates all triads in a pairwise comparison matrix, and finds the most inconsistent triad, the result does not depend on the order of alternatives.

*Proposition 14*. *KII* satisfies Axiom 3.
*Proof:* Let $b>1$ and let $x \equiv \frac{a_{ik}}{a_{ij}a_{jk}} > 1$ (without the loss of generality) be the most inconsistent triad. Then $KII = \min(1-\frac{1}{x}, x-1) = 1 - \frac{1}{x}$. Now $x^b \equiv \left(\frac{a_{ik}}{a_{ij}a_{jk}}\right)^b > 1$ and $KII^{(b)} = 1 - \frac{1}{x^b}$. Because $b>1$, so $\frac{1}{x} > \frac{1}{x^b}$ and $KII^{(b)} > KII$ as required.

*Proposition 15*. *KII* satisfies Axiom 4.
*Proof:* Let $x \equiv \frac{a_{ik}}{a_{ij}a_{jk}}$ denote the triad which is changed from a consistent state to an inconsistent one. In the consistent case $x \equiv \frac{a_{ik}}{a_{ij}a_{jk}} = 1$, otherwise $x \equiv \frac{a_{ik}}{a_{ij}a_{jk}} < 1$ or

$x \equiv \frac{a_{ik}}{a_{ij}a_{jk}} > 1$. Let us assume (without loss of generality) that $x \equiv \frac{a_{ik}}{a_{ij}a_{jk}} > 1$. Then, (10) reduces to finding the minimum of $\left\{1 - \frac{1}{x}, x - 1\right\}$. As $1 - \frac{1}{x} < x - 1$ for $x > 1$, $KII = 1 - \frac{1}{x}$ and $(KII)' = \frac{1}{x^2} > 0$; thus $KII$ is an increasing (monotone) function of $x \equiv \frac{a_{ik}}{a_{ij}a_{jk}}$. The more the triad goes inconsistent, the larger is $KII$ as required.

*Proposition 16*. $KII$ satisfies Axiom 5.
*Proof:* It is obvious from the definition of $KII$.

*Proposition 17*. $KII$ satisfies Axiom 6.
*Proof*: From the definition of $KII$ it is clear $KII$ is bounded from above by the value of 1.

Further, the indices introduced in the previous sections will be examined whether they satisfy the additional Axiom 6.

*Proposition 18*. C.I. does not satisfy Axiom 6.
*Proof*: Consider the corner pairwise comparison matrix of the order $n$. The characteristic polynomial of the matrix is: $\lambda_{max}^3 - n\lambda_{max}^2 = (n-2)(x + x^{-1} - 2)$, see Koczkodaj and Szwarc (2014).
Since $n \in N$ and $\lambda_{max} \geq n$, we have: $\lambda_{max}^3 > \lambda_{max}^3 - n\lambda_{max}^2 = (n-2)(x + x^{-1} - 2)$,
and $\lambda_{max} > \left[(n-2)(x + x^{-1} - 2)\right]^{1/3}$. From the last inequality for $n$ fixed and for $x \to +\infty$ we obtain $\lambda_{max} \to +\infty$, therefore C.I. is not bounded from above.

*Proposition 19*. GWI satisfies Axiom 6.
*Proof*: In GWI definition $\overline{A}(\overline{a}_{ij})$ is a normalized matrix with the sum of each column entries equal to 1. Let $K = \frac{1}{n}\sum_{i=1}^{n}\sum_{j=1}^{n}\overline{a}_{ij}$. Certainly, $K > GWI$ because $\overline{w}_i \geq 0, \forall i \in \{1,2,...,n\}$. Now, $K$ is the sum of $n$ rows, each equal to one, hence $K = 1$ for all $n$. Therefore, $K$ provides an upper bound for GWI as required.

*Proposition 20*. PLI does not satisfy Axiom 6.
*Proof* by counterexample: consider the corner pairwise comparison matrix of the order $n = 3$. Then, $PLI = x + 1/x$, which is not bounded from above.

*Proposition 21*. GCI does not satisfy Axiom 6.
*Proof* by counterexample: Consider the corner pairwise comparison matrix of the order $n = 3$. In this case we get: $GCI = \ln^2\left(\frac{w_2}{w_1}\right) + \ln^2\left(x\frac{w_3}{w_1}\right) + \ln^2\left(\frac{w_3}{w_2}\right)$. The first and the third terms are constant, but the middle one is logarithmically increasing, and because the logarithmic function is not bounded from above, GCI is not bounded from above as well.

# 6 Numerical example: a corner pairwise comparison matrix

To numerically illustrate behavior of selected indices, the corner pairwise comparison matrix $C$ (see below) of the order $n = 3$ will be examined. In the matrix $C$ all entries except for $a_{1n} \equiv x$ and $a_{n1} \equiv 1/x$ are set to 1, and $x \in R, x \geq 1$.

$$C = \begin{pmatrix} 1 & 1 & x \\ 1 & 1 & 1 \\ 1/x & 1 & 1 \end{pmatrix}$$

The dependence of selected inconsistency indices on $x$ is shown in Table 2, and Figure 1 and Figure 2. It should be noted that in this case $PLI = x + 1/x$.

As can be seen from this comparison, *RIC* is the least increasing index, while *PLI* is the most increasing one. *GCI*, *PLI*, and *C.I.* grow almost linearly. *KII* index grows rapidly for small $x$ (lower than 5), and then levels off.

**Table 2**. Comparison of inconsistency indices for different values of $x$.

| x | 1 | 2 | 3 | 4 | 5 | 6 | 7 | 8 | 9 | 10 | 100 |
|---|---|---|---|---|---|---|---|---|---|---|---|
| RIC | 0 | 0.0474 | 0.1011 | 0.1391 | 0.1658 | 0.1853 | 0.1999 | 0.2113 | 0.2204 | 0.2279 | 0.292 |
| CI | 0 | 0.027 | 0.068 | 0.109 | 0.147 | 0.184 | 0.218 | 0.25 | 0.28 | 0.309 | 1.428 |
| GWI | 0 | 0.1595 | 0.2509 | 0.3113 | 0.3547 | 0.3875 | 0.4134 | 0.4344 | 0.4518 | 0.4666 | 0.6492 |
| PLI | 0 | 0.5 | 1.3333 | 2.25 | 3.2 | 4.1667 | 5.1429 | 6.125 | 7.1111 | 8.1 | 98.1 |
| KII | 0 | 0.5 | 0.667 | 0.75 | 0.8 | 0.8333 | 0.857 | 0.875 | 0.889 | 0.9 | 0.99 |
| GCI | 0 | 0.1602 | 0.4023 | 0.6406 | 0.8634 | 1.07 | 1.2622 | 1.4414 | 1.6093 | 1.7676 | 7.0692 |

Source: author.

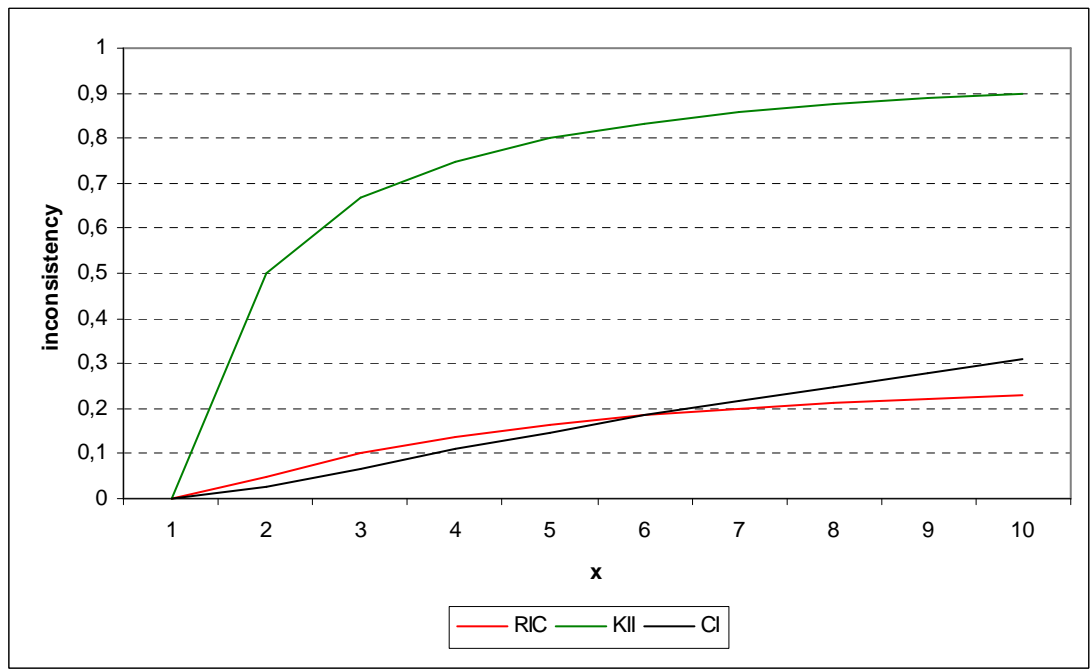

**Fig. 1.** A graphical comparison of inconsistency indices *RIC*, *KII* and *C.I.*
Source: author.

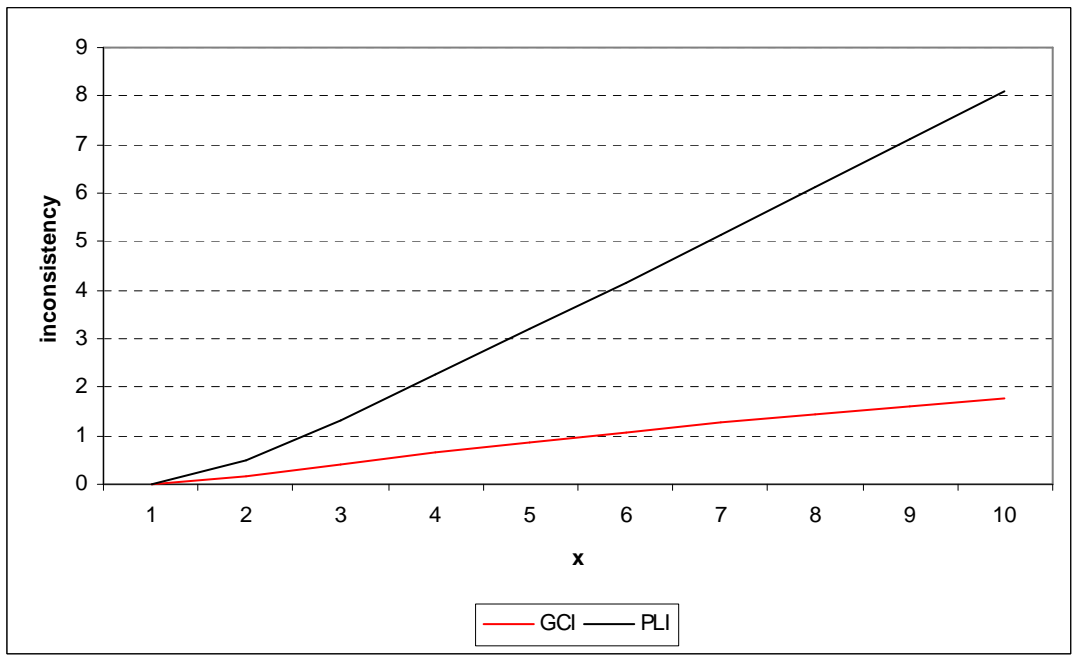

**Fig. 2.** A graphical comparison of inconsistency indices *GCI* and *PLI*.
Source: author.

## Conclusions

The aim of the paper was to introduce a new measure (*RIC*) of inconsistency of a pairwise comparisons matrix based on the dot product of its row vectors.
Furthermore, to the set of five axioms for inconsistency indices proposed by Brunelli and Fedrizzi (2015), an additional sixth axiom (upper boundary of index's values) was added.
New *RIC* index and Koczkodaj's index (*KII*) were examined with respect to these 6 axioms. Also, a (dis)satisfaction of the sixth axiom was proved for other indices, namely Saaty's *C.I.*, Golden-Wang's Index, Geometric Consistency Index and Peláez-Lamata's Index. Koczkodaj's index was the only index satisfying all six axioms.
Further study may focus on potential similarities among indices; see a comparative numerical study Brunelli et al. (2013), or an extension towards the fuzzy sets or the intuitionistic fuzzy sets. Also, theoretical studies on axiomatic systems for inconsistency indices would be of a considerable importance, as the axiomatic system proposed by Brunelli and Fedrizzi (2015) is likely not the optimal one yet.